\documentclass{article}

\usepackage[preprint]{neurips_2025}

\usepackage[utf8]{inputenc} 
\usepackage[T1]{fontenc}    
\usepackage{hyperref}       
\usepackage{url}            
\usepackage{booktabs}       
\usepackage{amsfonts}       
\usepackage{nicefrac}       
\usepackage{microtype}      
\usepackage{xcolor}         
\usepackage{amsmath}
\usepackage{graphicx}
\usepackage{enumitem}
\usepackage{fontawesome}
\usepackage{cleveref}

\definecolor{mydarkblue}{rgb}{0,0.08,0.45}
\definecolor{mydarkgreen}{RGB}{0, 139, 69}
\definecolor{MAEblue}{HTML}{C3DDEF}

\definecolor{darkblue}{rgb}{0, 0, 0.5}
\hypersetup{colorlinks=true, citecolor=darkblue, linkcolor=darkblue, urlcolor=darkblue}

\Crefname{theorem}{Theorem}{Theorems}
\Crefname{corollary}{Corollary}{Corollaries}
\Crefname{lemma}{Lemma}{Lemmas}
\Crefname{proposition}{Proposition}{Propositions}
\Crefname{claim}{Claim}{Claims}
\Crefname{remark}{Remark}{Remarks}
\Crefname{observation}{Observation}{Observations}
\Crefname{assumption}{Assumption}{Assumptions}
\Crefname{template}{Template}{Template}
\Crefname{definition}{Definition}{Definitions}
\Crefname{algorithm}{Algorithm}{Algorithms}
\Crefname{table}{Table}{Tables}
\Crefname{property}{Property}{Properties}
\Crefname{line}{Line}{Lines}

\DeclareMathOperator*{\argmax}{arg\,max}  
\DeclareMathOperator*{\rolloutpi}{\textcolor{red}{\mu}}
\DeclareMathOperator*{\trainerpi}{\textcolor{blue}{\pi}}

\title{Defeating the Training-Inference Mismatch via FP16}

\vspace{-5cm}

\renewcommand\footnotemark{}
\author{%
  Penghui Qi\textsuperscript{*\texttt{$\dagger$1,2}}\thanks{\llap{$^*$}Core Contributors.}\thanks{\llap{$^\dagger$}Project Lead.}, Zichen Liu\textsuperscript{*\texttt{1,2}}, Xiangxin Zhou\textsuperscript{*\texttt{1}}, \\
  \textbf{Tianyu Pang}\textsuperscript{\texttt{1}}, \textbf{Chao Du}\textsuperscript{\texttt{1}}, \textbf{Wee Sun Lee}\textsuperscript{\texttt{2}}, \textbf{Min Lin}\textsuperscript{\texttt{1}} \\
  \textsuperscript{\texttt{1}}Sea AI Lab~~~
  \textsuperscript{\texttt{2}}National University of Singapore\\
  \texttt{\faGithub~\url{https://github.com/sail-sg/Precision-RL}}
}

\begin{document}

\maketitle

\vspace{-0.7cm}

\begin{abstract}

Reinforcement learning (RL) fine-tuning of large language models (LLMs) often suffers from instability due to the numerical mismatch between the training and inference policies. While prior work has attempted to mitigate this issue through algorithmic corrections or engineering alignments, we show that its root cause lies in the floating point precision itself. The widely adopted BF16, despite its large dynamic range, introduces large rounding errors that breaks the consistency between training and inference. In this work, we demonstrate that simply reverting to \textbf{FP16} effectively eliminates this mismatch. The change is simple, fully supported by modern frameworks with only a few lines of code change, and requires no modification to the model architecture or learning algorithm. Our results suggest that using FP16 uniformly yields more stable optimization, faster convergence, and stronger performance across diverse tasks, algorithms and frameworks. We hope these findings motivate a broader reconsideration of precision trade-offs in RL fine-tuning.
\looseness=-1
\end{abstract}

\begin{figure}[h]
    \centering
    \vspace{-0.2cm}
    \includegraphics[width=\linewidth]{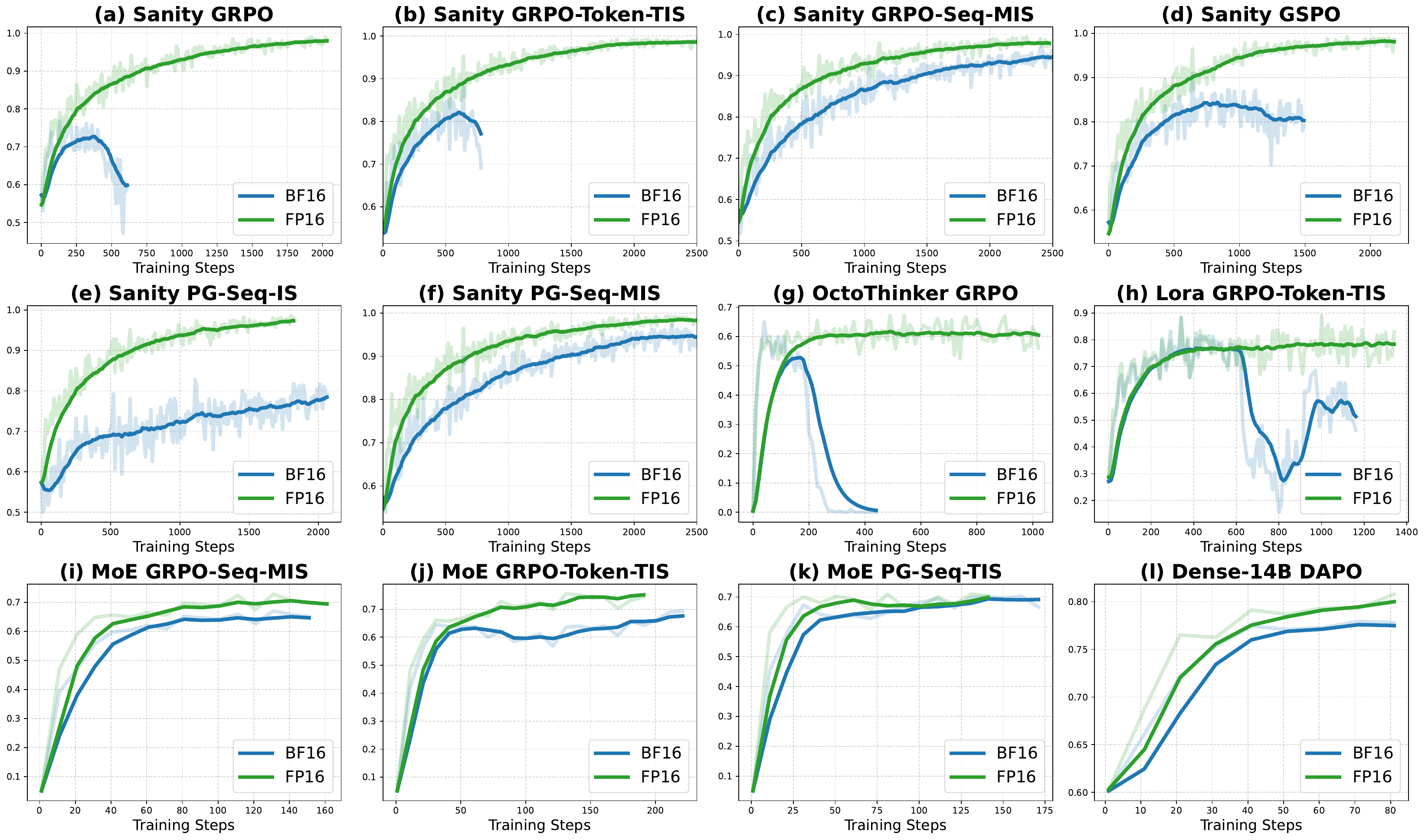}
    \caption{Training reward comparison between BF16 and FP16. We evaluate across diverse settings: our \texttt{Sanity} test (\Cref{sec:perfectible_rl}) with various algorithms (GRPO, GSPO, TIS, MIS, PG); different model families (R1D, Qwen and \texttt{OctoThinker}); alternative fine-tuning methods (\texttt{Lora}); and larger scale models (\texttt{Dense-14B}, \texttt{MoE}). Results are validated on two independent frameworks (VeRL and Oat).}
    \vspace{-0.3cm}
    \label{fig:bf16_vs_fp16_training}
\end{figure}

\section{Introduction}  
\label{sec:introduction}  






Reinforcement learning (RL) has emerged as a powerful paradigm for fine-tuning large language models (LLMs) to boost the reasoning performance~\citep{guo2025deepseekr1, zeng2025simplerl, liu2025understanding, qi2025optimizing}. However, the path to achieving high-performing models through RL is often fraught with instability. The training process is notoriously sensitive to hyperparameters and can suffer from training collapse, making it a significant challenge to reliably improve model performance~\citep{yao2025offpolicy, liu-li-2025, Team2025EveryAM, zheng2025group, yu2025dapo, cui2025entropy}. This fragility has spurred a continuous search for methods that can stabilize and streamline the RL fine-tuning process.  
  
A critical source of this instability stems from a fundamental discrepancy in modern RL frameworks: the \textbf{training-inference mismatch}. To accelerate training, these frameworks typically use different computational engines, a highly optimized one for fast inference (rollout) and another for training (gradient computation). While mathematically identical, these engines produce numerically different outputs due to precision errors and hardware-specific optimizations. As recent work has highlighted~\citep{yao2025offpolicy, liu-li-2025, Team2025EveryAM}, this seemingly minor mismatch between the inference and the training introduces significant issues into the optimization process.

Existing solutions have attempted to address this mismatch through algorithmic patches based on importance sampling. Notably, \citet{yao2025offpolicy} introduced a token-level importance sampling ratio as a patch to the GRPO~\citep{shao2024deepseekmath} gradient. While this simple correction can prolong training, it was later shown by \citet{liu-li-2025} to be insufficient to fully stabilize training due to its biased gradient. As an alternative, they proposed using an unbiased, sequence-level importance sampling ratio for the correction. Although this method is more stable, its effectiveness is hampered by slow convergence speed, a direct consequence of the high variance inherent in sequence-level ratios.  
Furthermore, both of these algorithmic approaches suffer from two fundamental problems:  
\begin{enumerate}  
    \item \textbf{They are computationally inefficient.} The implementations from \citet{yao2025offpolicy} and \citet{liu-li-2025} require an extra forward pass to compute the importance sampling ratio for their correction. Assuming a backward pass is twice the cost of a forward pass~\citep{qi2023zero}, this adds approximately 25\% to the training cost.  
    \item \textbf{The deployment gap persists.} By design, these solutions correct for the mismatch during training, but the final model parameters are optimized with respect to the training engine's probability distribution. This means the resulting model is not truly optimal for the inference engine used in deployment, which can lead to a tangible performance drop. This calls for a solution that eliminates the mismatch at its source, rather than merely compensating for it.  
\end{enumerate}

In this work, we take a step back from the complex algorithmic fixes and investigate the root cause of the numerical mismatch: \textbf{floating-point precision}. We identify that the modern standard for mixed-precision training, BFloat16 (BF16), is the primary culprit. While BF16 has a wide dynamic range which is excellent for stable pre-training, its low precision makes it highly susceptible to rounding errors that accumulate and eventually cause the training and inference policies to diverge.  
  
Our key finding is super simple: by switching from \textbf{BF16 to the FP16} during RL fine-tuning, we can virtually \textbf{eliminate the training-inference mismatch}. With more mantissa bits, FP16 offers higher numerical precision, making results less sensitive to the implementation differences between training and inference. The benefits of this simple change are multifold. It eliminates the complex algorithmic workarounds and the accompanying probability evaluations, restoring RL to its purest importance weighted policy-gradient form. It also closes the deployment gap that none of the existing fixes address. Empirical evaluations show a significant and uniform boost over both performance and stability, presenting a clean, efficient, and universally applicable solution to a critical challenge in RL-based LLM alignment.

\section{Background}

In modern RL frameworks for LLM fine-tuning, different engines are used for inference and training to maximize system efficiency, which inevitably creates a mismatch between the inference policy $\rolloutpi(\cdot|\theta)$ and training policy $\trainerpi(\cdot|\theta)$ due to subtle numerical discrepancies, even though, in principle, the two should be mathematically identical ($\rolloutpi=\trainerpi$). This mismatch brings two issues elaborated below,

\paragraph{Biased Gradient}

To optimize the trainer policy $\trainerpi(\cdot|\theta)$, we typically adopt the following objective:
\begin{equation}
    \mathcal{J}(\theta) = \mathbb{E}_{x \sim p_\mathcal{X}} \Bigl[ \mathcal{J}(x, \theta) \Bigr] = \mathbb{E}_{x \sim p_\mathcal{X}} \Bigl[ \mathbb{E}_{y \sim \trainerpi(\cdot|x, \theta)} [R(x, y) ] \Bigr],
\end{equation}
where $x$ is the prompt sampled from a distribution $p_\mathcal{X}$, $y$ is the response, and $R(x,y)$ is the reward of $y$.
The policy gradient can be calculated by REINFORCE estimator~\citep{williams1992simple, sutton2018rlbook}:
\begin{equation}
\begin{split}
    \nabla_\theta \mathcal{J}(\theta) &= \mathbb{E}_{x \sim p_\mathcal{X}} \Bigl[ \nabla_\theta \mathcal{J}(x, \theta) \Big], \\
    \nabla_\theta \mathcal{J}(x, \theta) 
    &= \mathbb{E}_{y \sim \trainerpi(\cdot|x, \theta)} \Bigl[\nabla_\theta \log \trainerpi(y|x, \theta) \cdot R(x, y) \Bigr].
\end{split}
\end{equation}
In practice, we sample the responses from the inference policy $\rolloutpi$, instead of the training policy $\trainerpi$. As noted by \citet{yao2025offpolicy} and \citet{liu-li-2025}, the policy gradient would become biased if simply ignoring this mismatch.
\begin{equation}
\begin{split}
    \nabla_\theta \mathcal{J}_\text{biased}(x, \theta) 
    &= \mathbb{E}_{y \sim \rolloutpi(\cdot|x, \theta)} \Bigl[ \nabla_\theta \log \trainerpi(y|x, \theta) \cdot R(x, y) \Bigr] \\
    &\neq \nabla_\theta \mathcal{J}(x, \theta) 
\end{split}
\end{equation}

\paragraph{Deployment Gap}

Another important but hard to fix issue is the deployment gap. Though it is $\trainerpi(\cdot|\theta)$ that we train, it is $\rolloutpi(\cdot|\theta)$ that we use for deployment and evaluation. However, the parameter $\theta$ optimized under the training engine $\trainerpi$ is not necessarily optimal for the inference engine $\rolloutpi$:
\begin{equation}
\begin{split}
    \argmax_\theta \mathbb{E}_{x \sim p_\mathcal{X}, y \sim \rolloutpi(\cdot|x, \theta)} [R(x, y) ] \neq \argmax_\theta \mathbb{E}_{x \sim p_\mathcal{X}, y \sim \trainerpi(\cdot|x, \theta)} [R(x, y)  ] 
\end{split}
\end{equation}
This deployment gap results in a non-trivial performance degrade due to this mismatch. While algorithmic patches~\citep{yao2025offpolicy, liu-li-2025} fix the biased gradient, by nature they cannot close the deployment gap, which calls for a fundamental solution to remove the mismatch altogether.

\subsection{Correcting Biased Gradient via Importance Sampling} \label{sec:algorithmic_fix}

To correct the biased gradient introduced by the training-inference mismatch, a principled approach is to use importance sampling (IS). This method re-weights the gradient calculation using a sequence-level probability ratio, ensuring the gradient estimator remains unbiased. The policy gradient for a given prompt $x$ is thus corrected as:
\begin{equation}
\label{eq:pg-seq-is}
\begin{split}
\nabla_\theta \mathcal{J}_\text{pg-is}(x)
&= \mathbb{E}_{y \sim \rolloutpi(\cdot|x, \theta')} \left[ \frac{ \trainerpi(y|x, \theta)}{\rolloutpi(y|x, \theta')} \nabla_\theta \log \trainerpi(y|x, \theta) \cdot A(x, y) \right],
\end{split}
\end{equation}
where $\theta'$ denotes the parameters used for sampling, which may differ from $\theta$ in an off-policy setting. The term $A(x, y) = R(x, y) - B(x)$ is the advantage, with $B(x)$ serving as a baseline for variance reduction~\citep{sutton2018rlbook}.

While theoretically sound, this estimator often suffers from high variance, particularly in the context of LLMs where response sequences are long, leading to extreme probability ratios. To mitigate this, techniques that trade a small amount of bias for a significant reduction in variance, such as Truncated Importance Sampling (TIS)~\citep{espeholt2018impala,yao2025offpolicy} and Masked Importance Sampling (MIS)~\citep{zheng2025group,team2025every,liu-li-2025}, have been proposed:
\begin{align}
\nabla_\theta \mathcal{J}_\text{pg-tis}(x) &= \mathbb{E}_{y \sim \rolloutpi(\cdot|x, \theta')} \left[ \min\left( \frac{\trainerpi(y|x, \theta)}{\rolloutpi(y|x, \theta')}, C \right) \cdot \nabla_\theta \log \trainerpi(y|x, \theta) \cdot A(x, y) \right], \label{eq:pg-seq-tis} \\
\nabla_\theta \mathcal{J}_\text{pg-mis}(x) &= \mathbb{E}_{y \sim \rolloutpi(\cdot|x, \theta')} \left[ \frac{\trainerpi(y|x, \theta)}{\rolloutpi(y|x, \theta')} \cdot \mathbb{I}\left\{ \frac{\trainerpi(y|x, \theta)}{\rolloutpi(y|x, \theta')} \le C \right\} \cdot \nabla_\theta \log \trainerpi(y|x, \theta) \cdot A(x, y) \right], \label{eq:pg-seq-mis}
\end{align}
where $C$ is a clipping hyperparameter and $\mathbb{I}\{\cdot\}$ is the indicator function. These methods stabilize training by controlling the magnitude of the importance weights.

\subsubsection{Existing Implementations}

Although generally inspired by the importance sampling principle, recent methods~\citep{yao2025offpolicy, liu-li-2025} are effectively implemented as auxiliary patches on top of GRPO, rather than adhering to the strictly principled formulation. Unfortunately, many widely used RL frameworks (e.g., VeRL~\citep{sheng2024hybridflow}) are GRPO-centric and do not natively provide the standard importance-weighted estimators outlined in \Cref{eq:pg-seq-is}, \Cref{eq:pg-seq-tis}, and \Cref{eq:pg-seq-mis}.

The standard GRPO gradient~\citep{shao2024deepseekmath, liu2025understanding}, which does not correct for the training-inference mismatch, is calculated as follows:\footnote{We use the Dr.GRPO variant to remove the length and difficulty biases of the vanilla GRPO.}
\begin{equation}
\label{eq:grpo}
\begin{split}
\nabla_\theta \mathcal{J}_\text{grpo}(x) &= \mathbb{E}_{y \sim \rolloutpi(\cdot|x, \theta')} \left[ \sum_{t=1}^{|y|} \nabla_\theta \min \left( r_t A_t, \text{clip}( r_t, 1-\epsilon, 1+\epsilon ) A_t \right) \right], \\
\text{where } r_t &= \frac{ \trainerpi(y_t|x, y_{<t}, \theta)}{\trainerpi(y_t|x, y_{<t}, \theta')} \text{ and } A_t = R(x,y) - \frac{1}{G-1}\sum_{i=1}^{G-1} R(x,y_i).
\end{split}
\end{equation}
For each prompt $x$, a group of $G$ responses $\{y_i\}_{i=1}^G$ is sampled from the inference policy $\rolloutpi(\cdot|x,\theta')$ to compute the advantage function $A_t$ as in GRPO and RLOO~\citep{ahmadian2024back, kool2019buy}.

Based on GRPO, \citet{yao2025offpolicy} introduced a token-level TIS correction:
\begin{equation}
\label{eq:grpo-tok-tis}
\footnotesize
\begin{split}
\nabla_\theta \mathcal{J}_\text{grpo-tok-tis}(x) &= \mathbb{E}_{y \sim \rolloutpi(\cdot|x, \theta')} \left[ \sum_{t=1}^{|y|} \min(\rho_t , C) \cdot \nabla_\theta \min \left( r_t A_t, \text{clip}( r_t, 1-\epsilon, 1+\epsilon ) A_t \right) \right], \\
\text{where } \rho_t &= \frac{ \trainerpi(y_t|x, y_{<t}, \theta')}{\rolloutpi(y_t|x, y_{<t}, \theta')}.
\end{split}
\end{equation}

Subsequently, \citet{liu-li-2025} advanced this approach by proposing a sequence-level MIS variant. This correction is applied to the entire GRPO gradient term, using a single ratio for the whole sequence to determine whether the update is applied:
\begin{equation}
\label{eq:grpo-seq-mis}
\footnotesize
\begin{split}
\nabla_\theta \mathcal{J}_\text{grpo-seq-mis}(x) &= \mathbb{E}_{y \sim \rolloutpi(\cdot|x, \theta')} \left[ \rho \cdot \mathbb{I}\{\rho \le C\} \cdot \sum_{t=1}^{|y|} \nabla_\theta \min \left( r_t A_t, \text{clip}( r_t, 1-\epsilon, 1+\epsilon ) A_t \right) \right], \\
\text{where } \rho &= \frac{ \trainerpi(y|x, \theta')}{\rolloutpi(y|x, \theta')}.
\end{split}
\end{equation}

Compared to the vanilla policy gradient estimators (\Cref{eq:pg-seq-is} and its TIS/MIS variants), existing GRPO-based implementations require an additional forward pass to compute $\trainerpi(\cdot|\theta')$ for their off-policy correction. This extra step incurs approximately 25\% computational overhead during training, assuming a backward pass is twice as costly as a forward pass~\citep{qi2023zero}.



\subsection{Engineering Attempts to Reduce the Mismatch}

Another line of work attempts to mitigate the training-inference mismatch from an engineering perspective, but with limited success. Early attempts, such as using an FP32 language model head by \citet{chen2025minimax}, is shown to be insufficient to prevent training collapse~\citep{yao2025offpolicy, liu-li-2025}. Very recently, \citet{Team2025EveryAM} reported promising results by manually aligning training and inference implementations. However, this approach requires deep domain knowledge and substantial engineering effort, and it is unclear whether such bespoke fixes can be generalized across different frameworks or models. A tangentially related work by \citet{he2025nondeterminism} demonstrated how to enforce determinism in inference, their method incurs a significant efficiency cost and cannot directly address the training-inference mismatch.

Despite these engineering efforts, the mismatch persists due to fundamental differences between training and inference computations that are difficult to reconcile. For example, tokens are generated auto-regressively during inference but are processed in parallel during training. Different parallelization strategies and precision-sensitive operations such as top-k expert selection in Mixture-of-Experts (MoE) models, further complicate the situation. This inherent difficulty highlights the need for a more fundamental solution that avoids such complex and brittle engineering workarounds.

\section{Revisiting FP16 Precision}
\label{sec:revisiting_fp16}  

In our investigation of the training–inference mismatch, we identify a surprisingly simple yet highly effective remedy that avoids complex algorithmic or engineering fixes. Rather than introducing additional machinery, we focus on a more fundamental factor: numerical precision. We find that merely switching the training precision from the now-dominant BF16 format~\citep{dean2012large, kalamkar2019study} to the earlier Float16 (FP16) format~\citep{micikevicius2017mixed} substantially mitigates the policy mismatch and yields significant performance improvements across RL algorithms. This section revisits the history and characteristics of these floating-point formats to shed light on this counterintuitive but powerful result.

\subsection{FP16 vs. BF16}  
Floating-point formats represent real numbers by dividing their bit budget between two components: exponent bits, which determine the range (how large or small a value can be), and mantissa bits (also known as fraction bits), which determine the precision (how finely values can be distinguished within that range). Both FP16 and BF16 use 16 bits in total, but they allocate these bits differently, resulting in distinct trade-offs between range and precision (see \Cref{tab:fp_comparison}).

\textbf{FP16} (IEEE 754 half-precision) allocates 5 bits to the exponent and 10 bits to the mantissa. The relatively large mantissa gives FP16 higher numerical precision, allowing it to represent small differences between nearby values accurately. However, its limited 5-bit exponent severely constrains the dynamic range, making FP16 prone to overflow (values exceeding the representable maximum) and underflow (values rounding to zero). Training with FP16 often requires stability techniques such as loss scaling to mitigate these issues (see \Cref{sec:loss_scaling}).

\textbf{BF16} (bfloat16), introduced by Google, allocates 8 bits to the exponent—matching the range of the 32-bit FP32 format—and only 7 bits to the mantissa. This design provides a wide dynamic range comparable to FP32, making BF16 highly resistant to overflow and underflow, at the cost of reduced precision. The resulting numerical robustness under low precision is the key reason for its widespread adoption in large-scale deep learning systems.

\begin{table}[h!]  
\centering  
\caption{Comparison of 16-bit Floating-Point Formats.}  
\label{tab:fp_comparison}  
\begin{tabular}{lcc}  
\toprule  
\multicolumn{1}{l}{\textbf{Property}} & \textbf{FP16} & \textbf{BF16} \\  
\midrule  
\textit{Bit Allocation} & & \\  
\quad Exponent Bits & 5 & 8 \\  
\quad Mantissa Bits & 10 & 7 \\  
\midrule  
\textit{Dynamic Range} & & \\  
\quad Smallest Positive Normal & $\approx 6.1 \times 10^\mathbf{-5}$ & $\approx 1.2 \times 10^\mathbf{-38}$ \\  
\quad Largest Value & $\approx 6.6 \times 10^\mathbf{4}$ & $\approx 3.4 \times 10^\mathbf{38}$ \\  
\midrule  
\textit{Precision} & & \\  
\quad Next Representable > 1 & $1 + 2^\mathbf{-10} \approx 1.000977$ & $1 + 2^\mathbf{-7} \approx 1.007812$ \\  
\bottomrule  
\end{tabular}  
\end{table}

\subsection{Stabilizing FP16 Training with Loss Scaling} \label{sec:loss_scaling}
The primary challenge with FP16's limited range is gradient underflow, which can be effectively solved early in the history of mixed-precision training with a technique called \textbf{loss scaling}~\citep{micikevicius2017mixed}. The procedure is straightforward:  
\begin{enumerate}  
    \item The loss is multiplied by a large scaling factor $S$ before backpropagation.  
    \item This scales up all gradients by $S$, shifting small gradient values out of the underflow region and into the representable range of FP16, thus preserving them.  
    \item Before updating the weights, the gradients are scaled back by dividing $S$.  
\end{enumerate}  
Modern implementations have further improved this with dynamic loss scaling. The scaling factor $S$ is automatically adjusted during training, increased if no overflows (infinity values in gradients) are detected for a number of steps, and decreased immediately if an overflow occurs.


Crucially, these loss scaling techniques are standard, mature components in mainstream training frameworks (e.g., PyTorch~\citep{paszke2019pytorch}, Megatron~\citep{shoeybi2019megatron}, DeepSpeed~\citep{rasley2020deepspeed}). Enabling them typically requires \textbf{only a single configuration change or a few lines of code}, making the adoption of FP16 training both simple and robust.

\subsection{The Rise of BF16 in Modern LLM Training}  

Despite the effectiveness of loss scaling, it complicates the system in distributed settings. Because a global synchronization is needed before the optimizer step to check for overflows and ensure the scaling factor is aligned across all workers. 

The introduction of BF16 on hardware like Google TPUs and later NVIDIA GPUs (starting with the Ampere architecture) is a game-changer. Having a same dynamic range as FP32, BF16 offered a ``drop-in'' replacement for FP32 that obviates meticulous loss scaling. Its resilience to overflow and underflow made training LLMs significantly more stable and straightforward. Consequently, BF16 quickly became the de-facto standard for modern mixed-precision training.

\subsection{Why FP16 is the Key for RL Fine-Tuning}  
While BF16's stability is an advantage for pre-training models, our findings reveal that its low precision is the origin of the training-inference mismatch.
  
Modern RL frameworks often use different engines or optimized kernels for training and inference. Even if both are configured to use BF16, subtle differences in their implementation (e.g., CUDA kernel optimizations, parallel strategies) can lead to different rounding errors on BF16. When these small discrepancies accumulate over a sequence of tokens during autoregressive sampling, the resulting probability distributions for $\trainerpi$ and $\rolloutpi$ can diverge significantly. This divergence is the source of the biased gradients and the deployment gap discussed earlier.  
  
This is precisely why switching to FP16 provides a fundamental solution. With its 10 mantissa bits, FP16 offers 8 times more precision ($2^{10}$ values vs. $2^7$ values) than BF16. This higher fidelity means that the outputs of the training and inference engines are much more likely to be numerically identical. The increased precision creates a buffer that absorbs the minor implementation differences between the two engines, preventing rounding errors from accumulating and causing a policy divergence.  

For RL fine-tuning, the dynamic range of the model's weights and activations has already been established during pre-training. Therefore, the extreme range of BF16 is less critical, while the precision it sacrifices becomes a dominant drawback. By reverting to FP16, we trade the unnecessary range of BF16 for the critical precision, effectively closing the gap between training and inference without any complex algorithmic or engineering workaround.

\subsection{Offline Analysis Results}

\begin{table}[]
    \centering
    \caption{Evaluation scores of DeepSeek-R1-Distill-Qwen-1.5B using under different precisions (BF16, FP16 and FP32) and token budgets (8K and 32K).}
    \begin{tabular}{ccccc}
        \toprule
        dtype & AMC23 (8K) & AIME24 (8K) & AMC23 (32K) & AIME24 (32K) \\
        \midrule
        BF16 & 50.38 & 22.60 & 62.35 & 29.90 \\
        FP16 & 50.60 & 20.10 & 63.10 & 30.94 \\
        FP32 & 51.54 & 22.30 & 62.42 & 28.44 \\
        \bottomrule
    \end{tabular}
    \label{tab:offline_eval_scores}
\end{table}

\begin{figure}[h]
    \centering
    \includegraphics[width=\linewidth]{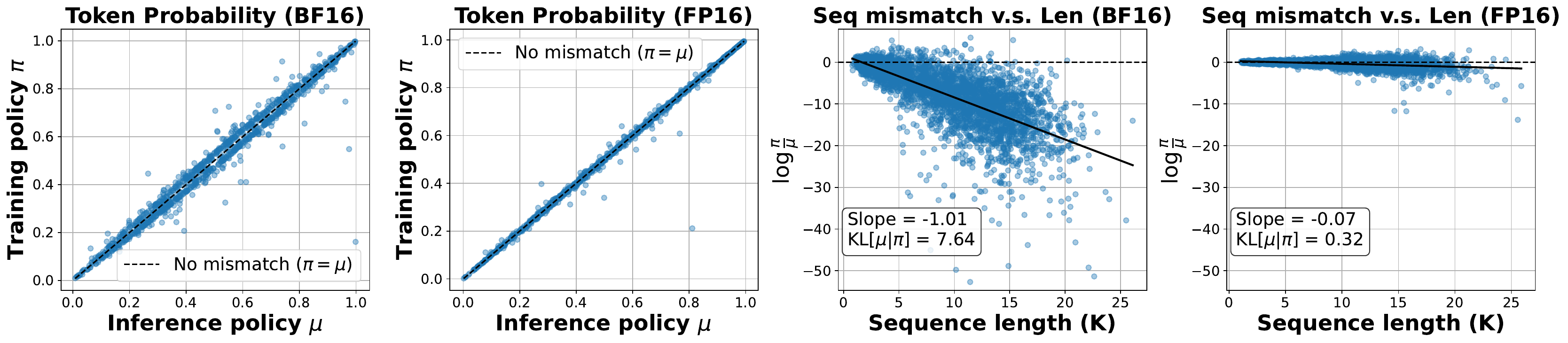}
    \caption{FP16 significantly reduces the training-inference mismatch. The left two plots show the token-level probability distribution, and the right two plots present the distribution of sequence-level log probability ratio between the inference policy ($\rolloutpi$) and the training policy ($\trainerpi$).
    Dashed lines in black denote perfect precision without mismatch.}
    \label{fig:offline_mismatch_32k}
\end{figure}

Before proceeding to RL fine-tuning, we first perform an offline analysis to examine performance and training–inference mismatch under different numeric precisions.
We begin by sampling 32 responses per question from the AMC and AIME benchmarks~\citep{li2024numinamath} using the DeepSeek-R1-Distill-Qwen-1.5B model\footnote{We follow their recommended decoding settings: temperature 0.6 and top-$p$ 0.95.}~\citep{guo2025deepseekr1}, with a 32K total token budget under both BF16 and FP16 precisions. As shown in \Cref{tab:offline_eval_scores}, their performance is largely comparable, suggesting that higher inference precision alone does not necessarily yield improvements.

Next, we re-generate 32 responses per question using temperature 1.0 and no top-$p$ sampling (so that $\rolloutpi$ is directly comparable to $\trainerpi$), and evaluate the token log-probabilities using the same model weights within the DeepSpeed training engine, under both BF16 and FP16 settings. The left two plots in \Cref{fig:offline_mismatch_32k} show the resulting distributions of token probabilities. We find that FP16 notably reduces the mismatch between $\rolloutpi$ and $\trainerpi$, with data points more tightly concentrated around the diagonal.

Beyond token-level discrepancies, we also analyze sequence-level mismatch, since $\frac{\trainerpi(y|x)}{\rolloutpi(y|x)}$ serves as an unbiased estimator of the importance sampling weight for a full response. The right two plots in \Cref{fig:offline_mismatch_32k} depict the distribution of sequence-level log-probability ratios across different generation lengths. The results clearly indicate that BF16 introduces an \textbf{exponentially larger mismatch}, which worsens with longer responses due to cumulative autoregressive errors, whereas FP16 maintains the mismatch at a much milder level (approximately 24$\times$ smaller).

\section{A Sanity Test for RL Algorithms}
\label{sec:perfectible_rl}

To rigorously assess the reliability and robustness of RL algorithms, we introduce a novel sanity test. Standard benchmarks often contain a mix of problems with varying difficulty, including questions that are either overly trivial or unsolvable by the initial model. Trivial questions waste computational resources, while unsolvable ones make it difficult to determine whether poor performance stems from a flawed algorithm or the model's inherent limitations. Our sanity test is designed to remove this ambiguity with efficiency. By creating a perfectible dataset where every problem is known to be solvable but not trivial, we can cleanly isolate and evaluate an RL algorithm's ability to unlock a model's latent potential. On this perfectible dataset, \textbf{a reliable RL algorithm should theoretically be able to achieve 100\% training accuracy}.

We construct this perfectible dataset by filtering out those overly trivial and unsolvable questions for the initial model. Specifically, we unroll 40 responses for each problem in the MATH dataset~\citep{hendrycks2021measuring}, and only keep problems where the initial accuracy is between 20\% and 80\%. This process yielded a targeted dataset of 1,460 questions for the DeepSeek-R1-Distill-Qwen-1.5B model~\citep{guo2025deepseekr1}. The smaller size of this dataset makes achieving near-100\% accuracy computationally feasible, allowing for efficient and conclusive testing.

We define our sanity test with a clear criterion: an RL algorithm passes if its training accuracy on this perfectible dataset converges above a high threshold (e.g., 95\%). An algorithm that fails this test can be considered unreliable or fundamentally flawed, as it is unable to guide the model to solve problems known to be within its reach. While passing is not a guarantee of universal success, \textbf{failing is a strong indicator of an ill-suited algorithm design}, making this test a crucial diagnostic tool.

\subsection{Experimental Setup}

Under this sanity test, we evaluate several representative RL algorithms, particularly those designed to address the training-inference mismatch (see \Cref{sec:algorithmic_fix}). All experiments use DeepSeek-R1-Distill-Qwen-1.5B as the initial model, with a context length of 8,000. We run each experiment on 8 NVIDIA A100 80G GPUs. For each policy iteration~\citep{schulman2017proximal}, we use a batch size of 64 questions (with 8 rollouts per question) and perform 4 gradient steps. For algorithms in the GRPO family, we set the \texttt{clip\_higher} to 0.28 by default~\citep{yu2025dapo}. The clipping threshold for importance sampling methods (\Cref{eq:pg-seq-mis} and \Cref{eq:grpo-seq-mis}) is set to $C=3$.


We evaluate a suite of methods designed to address the training-inference mismatch. This includes:  
\begin{itemize}  
    \item A vanilla GRPO baseline (specifically, the Dr.GRPO variant from \Cref{eq:grpo})~\citep{shao2024deepseekmath, liu2025understanding}.  
    \item GRPO with a token-level TIS correction (\Cref{eq:grpo-tok-tis}) from~\citet{yao2025offpolicy}.  
    \item GRPO with a sequence-level MIS correction (\Cref{eq:grpo-seq-mis}) from~\citet{liu-li-2025}.  
    \item The standard policy gradient algorithm with importance sampling (\Cref{eq:pg-seq-is}).  
\end{itemize}  
In addition, we include GSPO~\citep{zheng2025group} in our experiments, although it was primarily designed to address the mismatch introduced by MoE models.

\subsection{Comparison with Existing Algorithmic Corrections}
\label{sec:fp16_stabilizes}

\begin{figure}
    \centering
    \includegraphics[width=\linewidth]{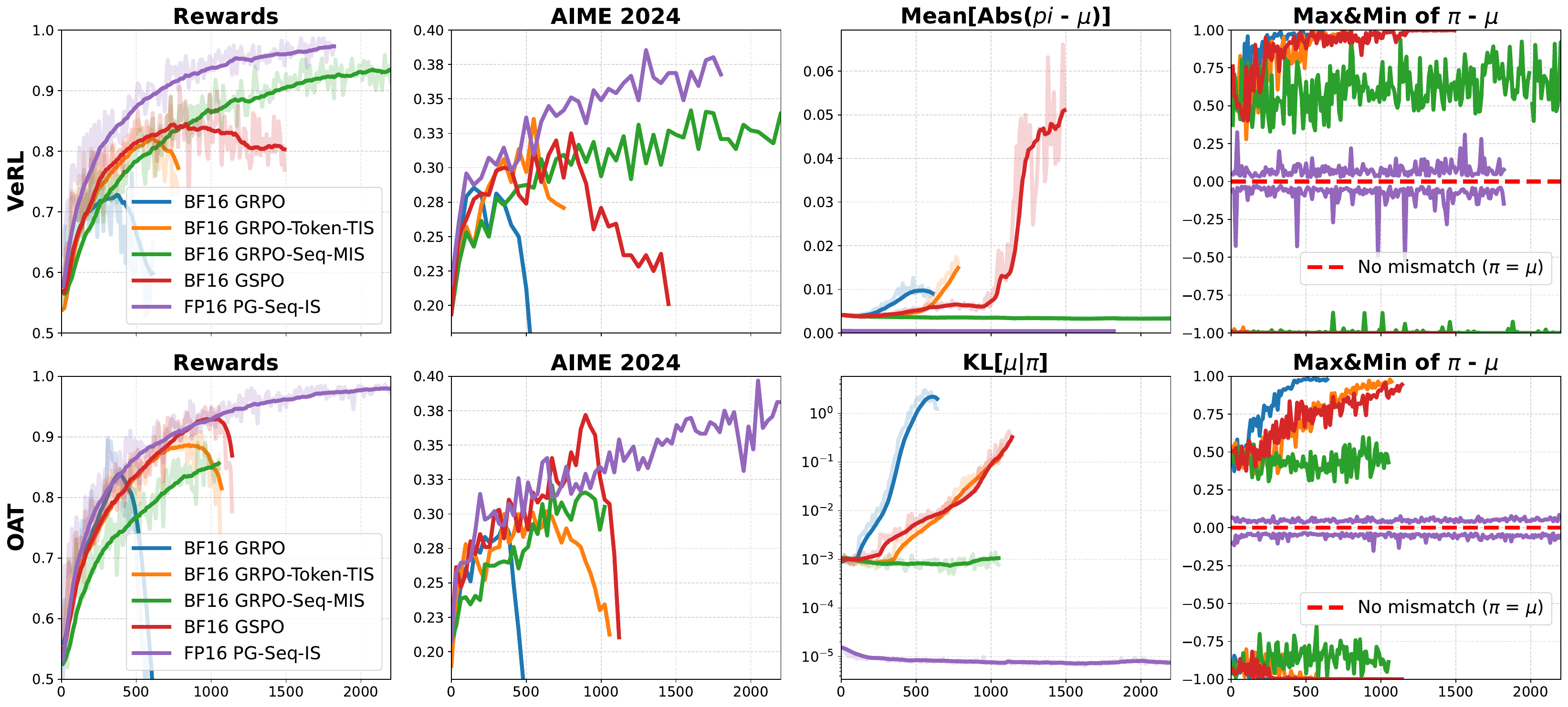}
    \caption{Simply switching from BF16 to FP16 stabilizes and prolongs RL training. The basic importance-weighted policy gradient algorithm in FP16 outperforms all baselines in BF16. Note that the third metric reported in each row slightly differs in implementation due to the use of separate codebases (VeRL and Oat). These metrics are semantically similar, and the minor differences do not affect our conclusions.}
    \label{fig:verl_oat_fix}
\end{figure}

To ensure robustness and rule out implementation-specific artifacts, we conducted experiments across two different frameworks: VeRL\footnote{We identified and corrected an implementation bug in VeRL's Dr.GRPO for our experiments. We optimized the training speed of VeRL based on \url{https://github.com/sail-sg/odc}.}~\citep{sheng2024hybridflow} and Oat~\citep{liu2025oat}. The results, shown in \Cref{fig:verl_oat_fix}, highlight the instability of existing methods when using BF16 precision.  

The vanilla GRPO baseline collapses early in training, reaching a peak accuracy of only 73\% in VeRL and 84\% in Oat before its performance degrades. The token-level TIS correction~\citep{yao2025offpolicy} prolongs training slightly but ultimately fails, collapsing after reaching 82\% (VeRL) and 88\% (Oat) accuracy, an observation that aligns with findings from~\citet{liu-li-2025}. Surprisingly, GSPO demonstrates more stable training for a longer period than GRPO with token-level TIS, achieving higher rewards despite not using the inference policy $\rolloutpi$ at all.\footnote{In our VeRL experiment, the GSPO gradient norm became `NaN' after 1200 steps, halting further model updates.}  
  
Among all the algorithmic corrections in BF16, only GRPO with sequence-level MIS~\citep{liu-li-2025} maintains stable training without collapsing. However, this stability is costly. The method suffers from slow convergence due to the high variance of its sequence-level importance ratio (see \Cref{fig:offline_mismatch_32k}). More importantly, even at its peak, it exhibits \textbf{a significant deployment gap} compared to our FP16 approach. It achieves a maximum training accuracy of only 95\% (vs. 99\% in FP16) and a score of 34\%  (vs. 39\% in FP16) on the AIME 2024 benchmark, demonstrating a clear performance ceiling. More evidence on deployment gap can be found in \Cref{fig:bf16_vs_fp16_training,fig:bf16_vs_fp16_eval}.

\paragraph{The Efficacy of FP16 Precision}  
  
In contrast to these algorithmic approaches, simply switching both training and inference precision from BF16 to FP16 provides a dramatic improvement. As shown in \Cref{fig:bf16_vs_fp16_training,fig:bf16_vs_fp16_eval}, the FP16 training runs are significantly more stable, converge much faster, and achieve substantially higher final rewards and evaluation scores across all tested algorithms. This result demonstrates that addressing the mismatch at the precision level is a more direct and effective solution than applying unstable or inefficient algorithmic corrections.

The most surprising finding is that FP16 precision fundamentally improves the behavior of importance sampling. The sequence-level ratio, which is notoriously high-variance, becomes much more concentrated and stable in FP16 (see \Cref{fig:offline_mismatch_32k}). This stabilization \textbf{makes it practical to use the classic, unbiased policy gradient estimator without any modifications} (\Cref{eq:pg-seq-is}). As shown in \Cref{fig:verl_oat_fix}, this simple, unbiased approach, when powered by FP16, dramatically outperforms all existing algorithmic corrections in BF16.

\paragraph{Training Dynamics} Our experimental results reveal an interesting phenomenon: algorithms that eventually collapse consistently exhibit a growing training-inference mismatch beforehand, making it a potential early-warning signal (see \Cref{fig:verl_oat_fix}). During this period, the policy difference $\trainerpi(\cdot|\theta') - \rolloutpi(\cdot|\theta')$ also converges to extreme values, where one policy's probability approaches 1 while the other's approaches 0, despite using the same copy of weights. We suspect this is driven by a particular optimization bias, though further validation is required. In contrast, stable algorithms maintain a bounded mismatch. Crucially, FP16 training shows a much lower mismatch level than any BF16 method. This inherent stability at the precision level explains why a simple policy gradient with FP16 can outperform all existing, more sophisticated solutions.

\paragraph{Framework-Specific Differences}
While our core conclusions hold across both the VeRL~\citep{sheng2024hybridflow} and Oat~\citep{liu2025oat} frameworks, we observed subtle implementation-dependent differences. Initially, the training-inference mismatch is slightly smaller in Oat than in VeRL; for example, the initial policy difference $\trainerpi(\cdot|\theta') - \rolloutpi(\cdot|\theta')$ has a minimum near -0.9 in Oat versus -1.0 in VeRL. Even under FP16, where both frameworks exhibit a small mismatch, VeRL was more prone to occasional numerical spikes. These subtle stability differences, which we attribute to their different distributed backends (DeepSpeed ZeRO vs. PyTorch FSDP), likely explain why Oat yields slightly higher training rewards, particularly for the algorithms that eventually collapse.

\subsection{Reviewing RL Algorithms under FP16}
\label{sec:algo_in_fp16}

\begin{figure}
    \centering
    \includegraphics[width=\linewidth]{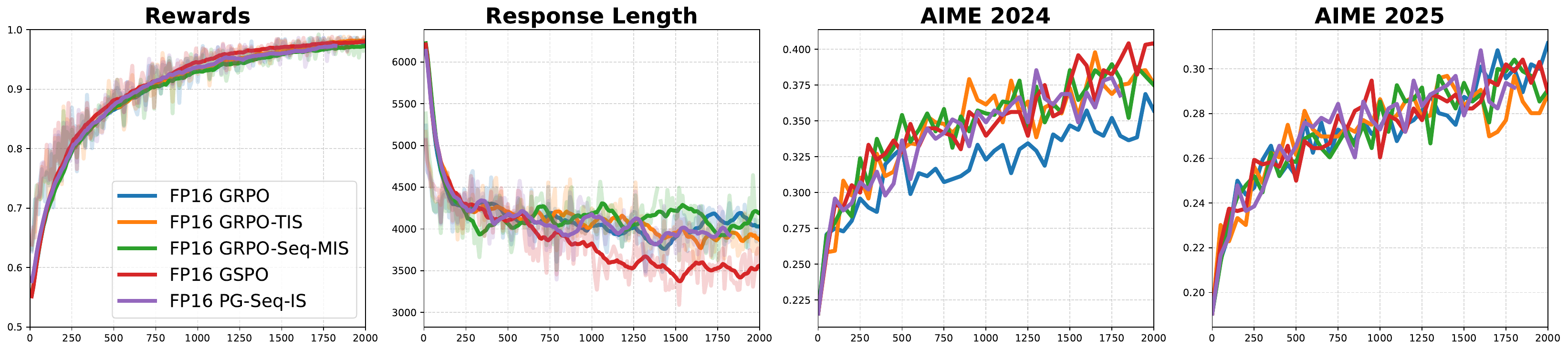}
    \caption{Comparisons between various algorithms based on FP16.}
    \label{fig:fp16_comparison}
\end{figure}

We then reviewed the performance of various RL algorithms when trained with FP16 precision. As shown in \Cref{fig:fp16_comparison}, the performance differences between algorithms become almost indistinguishable. We attribute this convergence in performance to the significantly reduced training-inference mismatch in FP16, which effectively transforms the optimization problem into a nearly on-policy setting. In this state, the complex corrections offered by different algorithms provide little to no additional benefit. We did observe a minor exception where the original GRPO scored slightly lower on the AIME 2024 benchmark; however, it also scored slightly higher on AIME 2025, making it difficult to draw a definitive conclusion about its relative performance.

\subsection{Ablation on the Precision}

\begin{figure}
    \centering
    \includegraphics[width=\linewidth]{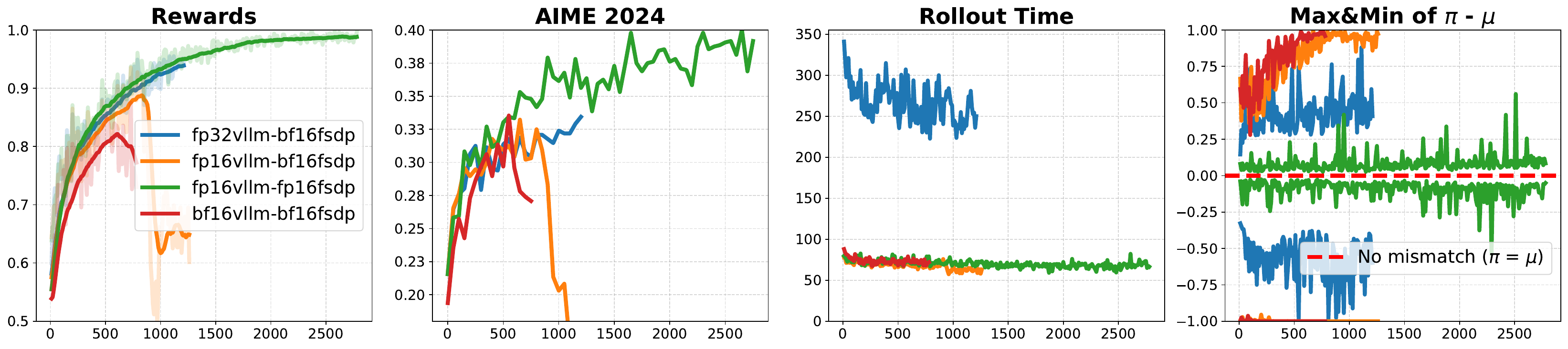}
    \caption{Ablation on the precision combinations.}
    \label{fig:precision}
\end{figure}

To isolate the effects of training and inference precision, we conducted an ablation study on the VeRL framework, using vLLM~\citep{kwon2023efficient} for inference and PyTorch FSDP~\citep{zhao2023pytorch} for training. The results are presented in Figure~\ref{fig:precision}.

When training with BF16 precision, we found that increasing the inference precision consistently prolonged training stability and improved performance. Notably, when paired with FP32 inference, the training run became fully stable with no signs of collapse. However, this stability came at an immense cost: FP32 inference was nearly three times slower than FP16 or BF16 inference, making this combination impractical for large-scale experiments.

In contrast, using FP16 for both training and inference yielded the best results. This combination not only produced the lowest training-inference mismatch but also resulted in the most stable training dynamics. It successfully reached nearly 100\% training accuracy on the perfectible dataset without any loss of speed, demonstrating a clear superiority in both stability and efficiency.

\section{Generalization Across Models, Data, and Training Regimes}

In \Cref{sec:perfectible_rl}, we scrutinized various algorithmic fixes under the sanity-check setting and found that simply switching from BF16 to FP16 can substantially improve training stability (\Cref{sec:fp16_stabilizes}), with its effect often overshadowing algorithmic tweaks (\Cref{sec:algo_in_fp16}).
In this section, we move beyond the sanity-check setting and validate our findings across more diverse scenarios, including Mixture-of-Experts (MoE) RL, Low-Rank Adaptation (LoRA) RL, and RL on larger prompt sets and alternative model families.

\subsection{MoE RL}
\label{subsec:moe_rl}

Mixture-of-Experts (MoE) reinforcement learning (RL) training is known for its instability and often requires sophisticated stabilization strategies~\citep{zheng2025group}.  
Both training and inference of MoE models typically involve distinct parallelization strategies and precision-sensitive operations such as top-$k$ expert selection, which further complicate the situation and usually lead to a larger training–inference mismatch compared to dense models.  
Given the widespread adoption of MoE architectures in modern LLMs, we conduct RL experiments on MoE models using Qwen3-30B-A3B-Base.  
We evaluate three different algorithms: GRPO-Seq-MIS, GRPO-Token-TIS, and PG-Seq-TIS, with detailed experimental settings provided in \Cref{appendix:moe_rl}.

Experiments using FP16 show greater stability and consistently higher training accuracies (see (i), (j), and (k) in \Cref{fig:bf16_vs_fp16_training}) as well as higher validation rewards (see (i), (j), and (k) in \Cref{fig:bf16_vs_fp16_eval}).  
The improvement is consistent across all three algorithms, indicating that adopting FP16 effectively mitigates the training–inference mismatch and enhances overall performance.

\subsection{LoRA RL}

LoRA~\citep{hu2022lora} has recently regained popularity in LLM RL~\citep{wang2025tina,schulman2025lora} due to its efficiency and performance comparable to full fine-tuning. To examine how LoRA-based RL is affected by numeric precision, we train Qwen2.5-Math-1.5B models on the standard MATH dataset using GRPO-Token-TIS (\Cref{eq:grpo-tok-tis}). LoRA is applied to all layers with a rank of $32$ and scaling factor $\alpha=64$. Following~\citet{schulman2025lora}, we adopt a slightly larger learning rate ($4\times10^{-5}$) than that used in full fine-tuning. As shown in \Cref{fig:bf16_vs_fp16_training} (h), BF16-based LoRA training collapses after roughly 600 steps, whereas FP16 maintains stable training throughout.

\subsection{RL on Large Dense Models}
\label{subsec:dense_rl}

Large-scale parameters are typically required in modern LLMs, yielding significantly better performance compared to smaller models. This motivates us to conduct RL experiments on large dense models. Specifically, we experiment with Qwen3-14B-Base and follow the algorithm of DAPO \citep{yu2025dapo}. Refer to \Cref{appendix:moe_rl} for details of experimental settings.

As shown in \Cref{fig:bf16_vs_fp16_training} (l), the training rewards with FP16 increase much faster than those with BF16. \Cref{fig:bf16_vs_fp16_eval} (l) demonstrates that FP16 achieves higher validation accuracy on AIME 2024. These results suggest that using FP16 instead of BF16 effectively mitigates the training–inference mismatch in large models, highlighting the potential of this approach for scaling RL training on large models.

\subsection{RL on Other Model Families}

The base models, which serve as the initial policies for RL, can substantially influence the learning dynamics, as they determine not only the scope of exploration but also the numerical range and sensitivity of network parameters and activations. To strengthen our experimental conclusions, we extend our study beyond Qwen-based models and train OctoThinker-3B~\citep{wang2025octothinker}, a model mid-trained from Llama3.2-3B~\citep{grattafiori2024llama} on reasoning-intensive data using GRPO. As shown in \Cref{fig:bf16_vs_fp16_training} (g), BF16 training destabilizes after around 150 steps due to numerical mismatch, while FP16 continues to train smoothly without collapse.

\section{Discussions}

\paragraph{Rethinking the Precision Tradeoff in RL Fine-Tuning}
Numerical precision is a foundational choice in the LLM training stack, yet this choice has long been dominated by BF16 for both pre-training and post-training, prized for its wide dynamic range and ease of use. Our results, however, suggest this default deserves careful rethinking for RL fine-tuning. In this phase, the training-inference mismatch becomes a critical source of instability, and BF16's low precision exacerbates this problem. We demonstrate that by simply trading BF16's wide dynamic range for FP16's higher precision, one can achieve significantly more stable RL training, faster convergence, and superior final performance.

It is important to note that we are not claiming FP16 is a universally optimal choice. The pursuit of efficiency may lead developer to even lower precisions like FP8. Furthermore, using FP16 for extremely large models might present engineering challenges related to its limited range, such as managing potential overflows. However, we believe these are solvable challenges, as evidenced by the recent successes in large-scale FP8 training. Ultimately, we hope this work inspires the community to reconsider FP16 as a powerful and often more suitable alternative for stabilizing RL fine-tuning.

\paragraph{The Bias-Variance Tradeoff under BF16 Precision}
Our results in \Cref{sec:fp16_stabilizes} reveal a bias-variance trade-off among RL algorithms operating under BF16 precision. Methods with lower variance but higher bias (like GRPO, token-level TIS, and GSPO) initially converge quickly but prove unstable and eventually collapse. Conversely, less biased algorithms that more accurately correct for the policy mismatch (like PG-Seq-IS and GRPO-Seq-MIS) achieve stability but at the cost of high variance, which slows their convergence.

This trade-off, however, becomes far less critical under FP16 precision. By fundamentally reducing the training-inference mismatch, FP16 naturally lowers both the bias induced by the mismatch and the variance of the importance sampling corrections. This enhanced stability allows even the most naive policy gradient estimator to converge efficiently, creating a training dynamic where all tested algorithms perform well and the tension between stability and speed is effectively resolved.

\section{Conclusion}

This work demonstrates that the training-inference mismatch, a major source of instability in RL fine-tuning, is fundamentally a problem of numerical precision. While existing algorithmic fixes are often complex and inefficient, we show that simply switching from the standard BF16 format to the higher-precision FP16 format can virtually eliminate the mismatch. This single, efficient change leads to more stable training, faster convergence, and superior performance, proving that addressing the problem at the precision level is a more effective strategy. We conclude that FP16 should be reconsidered as a foundational option for robust RL fine-tuning of LLM.

\bibliographystyle{plainnat}
\bibliography{reference}


\clearpage
\appendix

\section{Detailed Experimental Settings}
\label{sec:setup}

\subsection{MoE RL}
\label{appendix:moe_rl}

As for experiments of MoE RL, we use Qwen3-30B-A3B-Base as the base model. The training data comes from DAPO-Math-17k~\citep{yu2025dapo}, and we conduct online evaluation on AIME 2024 using the \texttt{avg@32} metric. The training is performed with the VeRL framework~\citep{sheng2024hybridflow}, and the key hyperparameters are summarized in \Cref{tab:exp_setting_moe_rl_and_dense_rl}. 

Dr.GRPO~\citep{liu2025understanding} proposes using a constant normalizer instead of a token-count-based normalizer. Notably, the open-source VeRL implementation does not correctly implement this. We refer to our corrected version as ``seq-mean-token-sum-norm'' for \texttt{actor.loss\_agg\_mode} in VeRL.

\subsection{RL on Large Dense Models}
\label{appendix:dense_rl}

For experiments on large dense models, we use Qwen2.5-14B-Base as our base model. The training data is sourced from the mathematical domain dataset curated by \citet{cheng2025revisiting}. They aggregated recent math reasoning collections including OR1~\citep{skywork-or1-2025}, DAPO~\citep{yu2025dapo}, and DeepScaler~\citep{deepscaler2025}, and then performed deduplication and filtering to derive a final collection of 54.4k math training samples. We conduct online evaluation on AIME 2024 using the \texttt{avg@8} metric. The training algorithms and hyperparameters follow the setup described in \citet{yu2025dapo}, as summarized in \Cref{tab:exp_setting_moe_rl_and_dense_rl}.

\begin{table}[h]
    \caption{Hyperparameters used for RL training of MoE models and large dense models.}
    \label{tab:exp_setting_moe_rl_and_dense_rl}
    \centering
    \begin{tabular}{lcc}
\toprule
\textbf{Parameter} & \textbf{MoE RL} & \textbf{Large dense RL}  \\
\midrule

\texttt{trainer.nnodes} & 8 & 8 \\
\texttt{trainer.n\_gpu\_per\_node} & 8 & 8 \\
\texttt{model.path} & Qwen3-30B-A3B-Base  & Qwen3-14B-Base \\
\texttt{vllm\_version}	& 0.10.0 & 0.10.0  \\
\texttt{data.train\_batch\_size}	& 512 & 512 \\
\texttt{data.gen\_batch\_size} & N/A & 1536 \\
\texttt{data.max\_prompt\_length} & 2048 & 2048 \\
\texttt{data.max\_response\_length} & 20480 & 20480 \\
\texttt{rollout.n} & 16 & 16 \\
\texttt{rollout.temperature} & 1.0& 1.0\\
\texttt{rollout.top\_p} & 1.0 & 1.0 \\
\texttt{val\_kwargs.temperature} & 0.6 & 1.0 \\
\texttt{val\_kwargs.top\_p} & 1.0 & 0.7\\
\texttt{actor.ppo\_mini\_batch\_size} & 32 & 32 \\
\texttt{actor.ppo\_max\_token\_len\_per\_gpu} & 22528 & 22528 \\
\texttt{optim.lr} & 1e-6 & 1e-6 \\
\texttt{optim.lr\_warmup\_steps} & N/A & 10 \\
\texttt{optim.weight\_decay} & 0.0 & 0.1 \\
\texttt{optim.betas} & [0.9, 0.95] & [0.9, 0.999] \\
\texttt{optim.eps} & 1e-15 & 1e-8 \\
\texttt{algorithm.use\_kl\_in\_reward} & False & False \\
\texttt{actor.use\_kl\_loss} & False & False \\
\texttt{actor.clip\_ratio\_high} & 0.28 & 0.28 \\
\texttt{actor.clip\_ratio\_low} & 0.2 & 0.2 \\
\texttt{actor.clip\_ratio\_c} & N/A & 10.0 \\
$C$ in \Cref{eq:pg-seq-tis,eq:pg-seq-mis} & 3.0 & N/A \\
\texttt{actor.loss\_agg\_mode} & seq-mean-token-sum-norm & token-mean \\
\texttt{overlong\_buffer.enable} & False & True \\
\texttt{overlong\_buffer.len} & N/A & 4096 \\
\texttt{overlong\_buffer.penalty\_factor} & N/A & 1.0 \\
\texttt{filter\_groups.enable} & False & True \\
\texttt{filter\_groups.metric} & N/A & acc \\
\texttt{filter\_groups.max\_num\_gen\_batches} & N/A & 10 \\
\bottomrule
\end{tabular}
\end{table}

\section{More Experimental Results}

\begin{figure}[h]
    \centering
    \includegraphics[width=\linewidth]{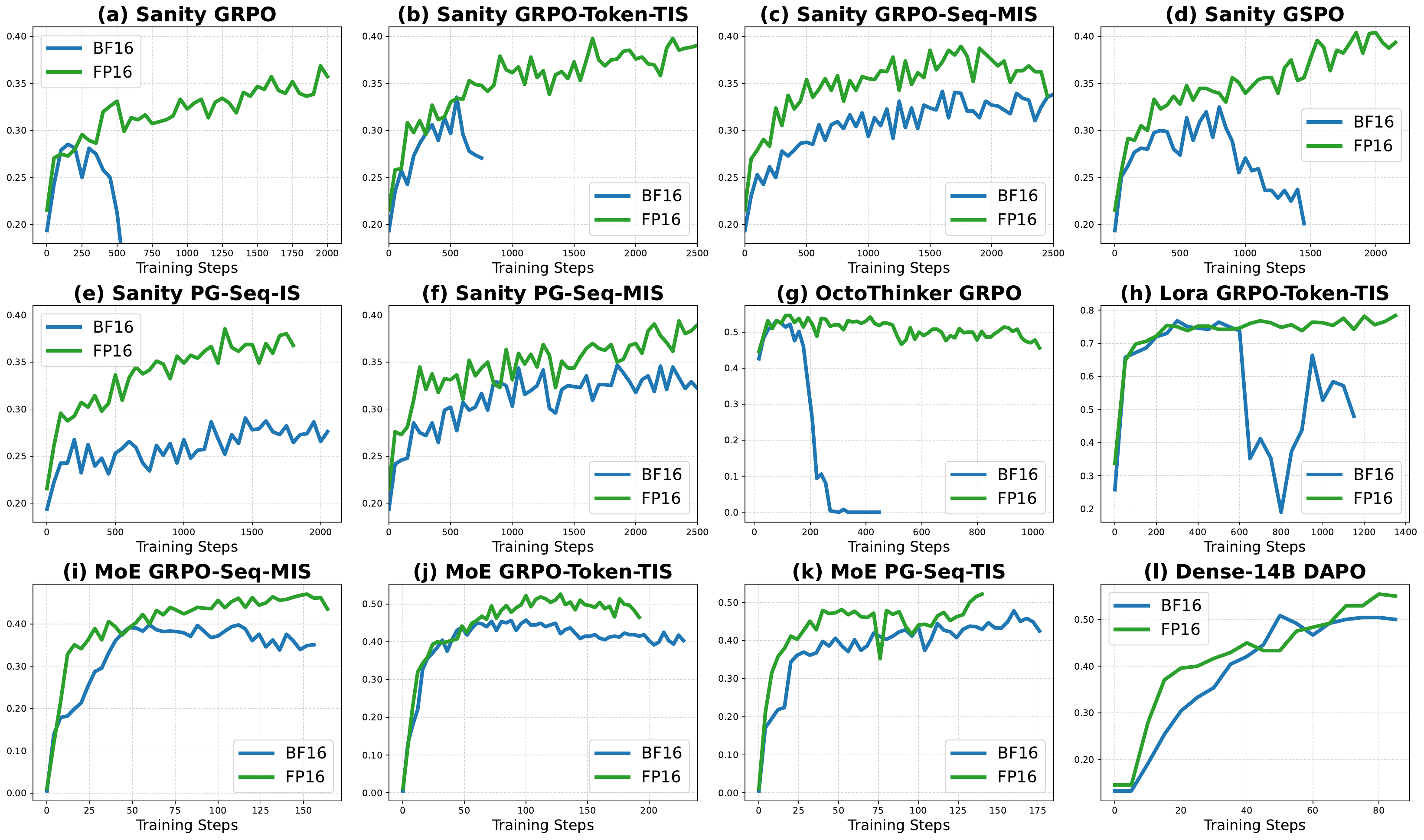}
    \caption{Evaluation comparisons between BF16 and FP16 across various frameworks, algorithms, datasets and training regimes.}
    \label{fig:bf16_vs_fp16_eval}
\end{figure}

While \Cref{fig:bf16_vs_fp16_training} presents the training reward curves under different precisions, \Cref{fig:bf16_vs_fp16_eval} shows evaluation results using checkpoints trained with these precisions. The results indicate that FP16-trained models generalize well to unseen benchmarks, further supporting our claim.


\end{document}